\documentclass[3p,times,procedia]{elsarticle}
\usepackage{ecrc}
\usepackage[bookmarks=false]{hyperref}
    \hypersetup{colorlinks,
      linkcolor=blue,
      citecolor=blue,
      urlcolor=blue}

\volume{00}

\firstpage{1}

\journalname{Procedia Computer Science}

\runauth{Author name}

\jid{procs}

\usepackage{amssymb}

\usepackage{algorithm}       
\usepackage{algpseudocode}   
\usepackage{amsmath}         
\usepackage{multirow}
\usepackage{booktabs}
\usepackage{graphicx}
\usepackage{subcaption}
\usepackage{placeins}
\usepackage[compact]{titlesec}

\titlespacing*{\section}{0pt}{1.0ex plus .2ex minus .2ex}{0.8ex plus .2ex}

\usepackage[font=small,skip=3pt]{caption}          

\newcommand{\figref}[1]{Fig.~\ref{#1}}
\newcommand{\tabref}[1]{Table~\ref{#1}}

\usepackage[figuresright]{rotating}

\begin{document}
\begin{frontmatter}



\dochead{7th International Conference on Industry of the Future and Smart Manufacturing}

\title{From Prompt to Graph: Comparing LLM-Based Information Extraction Strategies in Domain-Specific Ontology Development}


\author[a]{Xuan Liu\thanks{Xuan Liu and Ziyu Li contributed equally to this work.}} 
\author[b,f]{Ziyu Li\footnotemark[1]} 
\author[a]{Mu He}
\author[c]{Ziyang Ma}
\author[e]{Xiaoxu Wu}
\author[e]{Gizem Yilmaz}
\author[e]{Yiyuan Xia}
\author[d]{Bingbing Li\corref{cor1}}
\author[e]{He Tan}
\author[a]{Jerry Ying Hsi Fuh}
\author[a]{Wen Feng Lu}
\author[f]{Anders E.W. Jarfors\corref{cor2}}
\author[b]{Per Jansson}
\address[a]{Department of Mechanical Engineering, National University of Singapore, 9 Engineering Drive 1, 117575, Singapore}
\address[b]{Comptech i Skillingaryd AB, P.O. Box 28, 568 31 Skillingaryd, Sweden.}
\address[c]{School of Energy Systems, Lappeenranta University of Technology, Mukkulankatu 19, FI-15210 Lahti, Finland}
\address[d]{Department of Manufacturing Systems Engineering and Management, California State University Northridge, 18111 Nordhoff St, California 91330, U.S.A.}
\address[e]{Department of Computing, School of Engineering, Jönköping University, P.O. Box 1026, 551 11 Jönköping, Sweden.}
\address[f]{Department of Materials and Manufacturing,
School of Engineering, Jönköping University, P.O. Box 1026, 551 11 Jönköping, Sweden.}

\begin{abstract}
Ontologies are essential for structuring domain knowledge, improving accessibility, sharing, and reuse. However, traditional ontology construction relies on manual annotation and conventional natural language processing (NLP) techniques, making the process labour-intensive and costly, especially in specialised fields like casting manufacturing. The rise of Large Language Models (LLMs) offers new possibilities for automating knowledge extraction. This study investigates three LLM-based approaches, including pre-trained LLM-driven method, in-context learning (ICL) method and fine-tuning method to extract terms and relations from domain-specific texts using limited data. We compare their performances and use the best-performing method to build a casting ontology that validated by domian expert.
\end{abstract}


\end{frontmatter}

\email{bingbing.li@csun.edu,anders.jarfors@ju.se}



\section{Introduction and literature review}
\label{intro}
In the context of Industry 4.0, knowledge management (KM) techniques serve the entire lifecycle of group and organisational knowledge, from creation and storage to transfer and application. These techniques are recognised for their substantial contributions to various corporate facets, encompassing digital transformation, innovation ecosystems, decision-making, and production optimization \cite{manesh2020knowledge}. Deploying a KM system mitigates the poor generalization problem of data-driven models in handling new domains \cite{jawad2023adoption}.

Among KM techniques, ontology has emerged as a powerful tool for structuring domain knowledge in formal, machine-readable formats  \cite{inbook,Osman2022}. 
By defining concepts, relationships, and rules, ontologies provide a shared semantic foundation that facilitates interoperability, consistent data interpretation, and logical reasoning\cite{Arp2018}. Ontology-based approaches have proven especially effective in domains such as life sciences, material science, and engineering, enabling explicit representation of complex domain specific knowledge.

The manufacturing sector has similarly benefited from ontological approaches by enabling standardized data sharing, improving knowledge management, and enhancing decision-making across complex industrial environments. Ontological approaches can support production scheduling, operational control, and product design by providing structured vocabularies that facilitate cross-disciplinary collaboration and capture both explicit and implicit knowledge \cite{Borgo2007,senthil,Grny2014,Costa2013}. In engineering design, ontologies aid in organizing functional knowledge, improving classification, and reducing design iterations \cite{Cao2022,Bittner2005,Kitamura2003}. Additionally, in distributed or cloud-based systems, ontologies help manage heterogeneous data and encode expert knowledge into interpretable formats \cite{Sarkar2019,MartinezLastra2008}. This fosters efficiency, consistency, and reliability across the manufacturing lifecycle. 


However, a major bottleneck remains: the high cost and time intensity of ontology construction. Traditional ontology development relies heavily on manual annotation, which is especially problematic in complex industrial domains like manufacturing, where documentation is often fragmented and unstructured \cite{Rani2017,datatrasferforcasting,diecastingtrasnfer}.
To reduce manual effort, ontology learning has gained attention, involving steps such as term extraction, relation identification, and concept hierarchy building \cite{Du2024}. Recent advances in LLMs offer promising new directions in automating these tasks \cite{Joachimiak2024,Zhang2024,tan2024,Liao2025}. Funk et al. \cite{Funk2023} demonstrated that while general-purpose LLMs support hierarchical structuring, fine-tuned models yield better results in specialised domains. Liu et al. \cite{Liu2025} further showed that guided LLM annotation significantly improves the efficiency and accuracy of term and relation extraction.

Despite promising progress, existing studies rarely conduct side-by-side comparisons of different LLM-based ontology learning approaches. Furthermore, the manufacturing ontologies developed by most studies are informal, without complete concept-relation triples, limiting their practical utility. In addition, most prior research assumed access to large, well-structured datasets, which are not typical condition in traditional manufacturing settings, where documentation is mostly fragmented and unstructured.

To address these challenges, we present a semi-automated ontology learning framework that leverages LLMs to facilitate domain-specific term extraction and relation identification. In this study, casting serves as a representative case to evaluate the effectiveness of three LLM-based methods: pre-trained LLMs with optimised prompting, few-shot in-context learning (ICL), and fine-tuned models. The extracted knowledge is structured as triples—(\textit{subject, object, relation})—a graph-based expression format. These triples are used to build and implement an ontology in Neo4j, a popular graph database, using Cypher queries that are automatically generated by LLMs to construct the graph. 

The main contributions of this work are as follows: (1) a comparative evaluation of three LLM-based methods for term extraction and relation identification in domain-specific settings; (2) the proposal of a prompt-based relation extraction method designed to improve ontology construction in manufacturing applications; and (3) a well-constructed and evaluated ontology of casting.  
constructed from high-quality academic casting literature and technical books and validated by casting domain experts.

This work has been organised as follows: The first section describes the research background and also the related literature, as well as the current gap existing in the field. Section 2 describes different proposed methods and related experimental processes. Section 3 evaluates, discusses, and compares the experimental results. The last section concludes the paper and outlines directions for future work.

\section{Experiment and Methods}
In this study, we evaluate and compare three LLM-assisted information extraction approaches for ontology construction: (1) extraction supported by a general pre-trained LLM, (2) ICL-based extraction, and (3) extraction using fine-tuned LLMs incorporating domain-specific knowledge. Among them, the ICL method and the fine-tuning method extract information triples in two steps, extracting both terms and relations with well-designed prompts.

\begin{figure}[!h]
    \centering
    \includegraphics[width=0.98\linewidth]{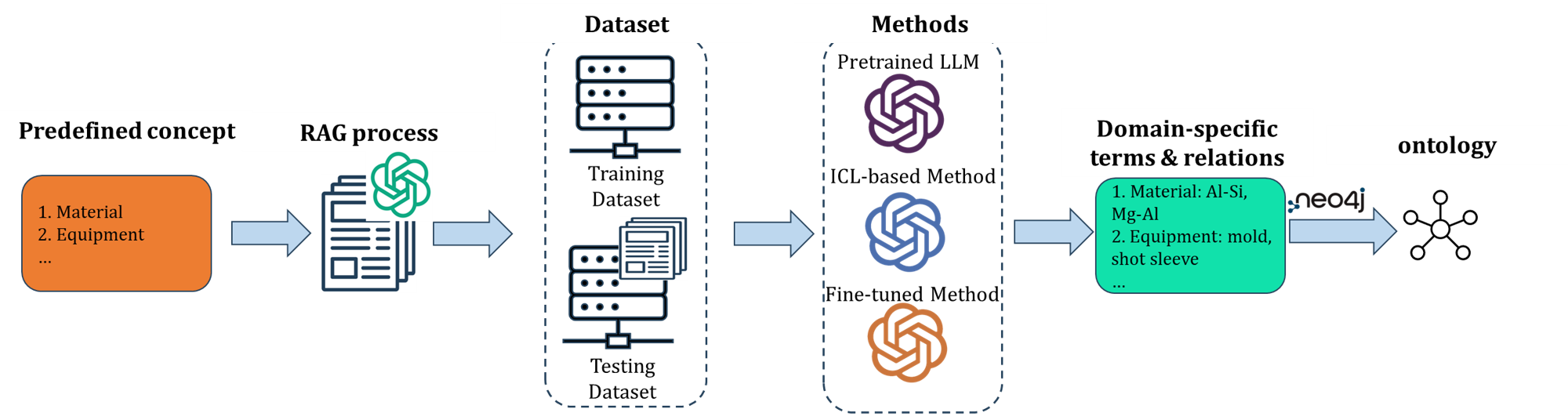}
    \caption{Overall experimental plan for ontology construction}
    \label{fig:1}
\end{figure}

Before extraction, to facilitate the ontology construction process, 6 top concepts (\texttt{materials, casting process, product property, casting parameter, casting defect, and casting equipment}) were identified by casting domain expert. These concepts guided the data collection and annotation and supported ontology construction. 
Detailed descriptions of the experimental procedures for each method are provided in the following sections. \figref{fig:1} illustrates the whole experimental process of this study.

\subsection{Data Preparation}

 The data preparation process encompassed multiple stages: text data collection, pre-processing, selection, and related terms and relation annotation. Text data relevant to the 6 topics was sourced from two primary repositories: academic papers and domain specific technical books. 

\subsubsection{Retrieval augmented generation (RAG) based data distillation process}

\begin{figure}[!h]
    \centering
    \includegraphics[width=0.98\linewidth]{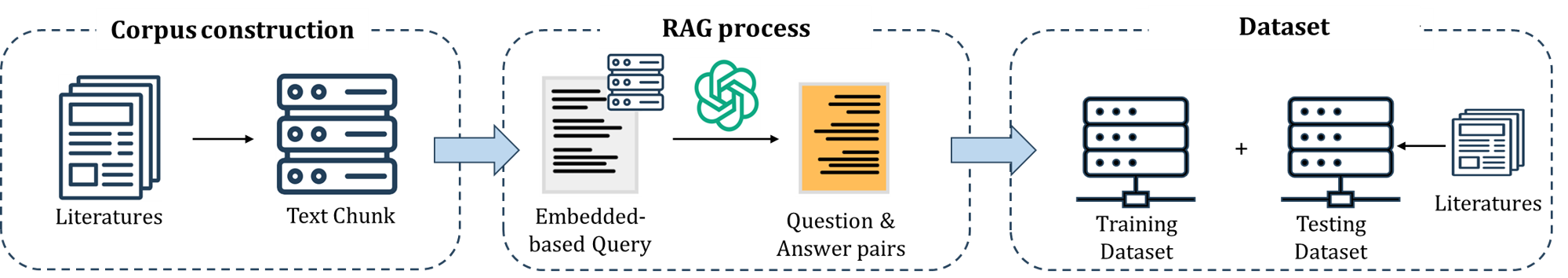}
    \caption{Knowledge distillation and dataset preparation}
    \label{fig:2}
\end{figure}

Following the initial data collection and pre-processing, the knowledge distillation method proposed by Fan et al.\cite{Fan2024Unleashing} was adopted. Using LLMs in combination with RAG techniques, illustrated in \figref{fig:2}, the process condensed extensive text into focused summaries while preserving essential domain knowledge. The final presentation in this section is question answer pairs. 

\subsubsection{Training and testing dataset preparation}
The distilled results underwent domain expert evaluation, after which a curated subset was selected to construct the training and testing datasets. Specifically, the training dataset consisted of 173 short texts, each containing 1-3 sentences derived from the validated retrieval output, with balanced topic representation (about 30 texts per topic). This dataset are used to identify the best-performed prompt in the pre-trained LLM-driven method, provide demonstrations for few-shot ICL method and fine-tune the domain-specific information extraction LLMs. For testing, 30 long paragraphs, each containing at least 6 sentences, were chosen. Among them, 20 were the texts from the question and answer pairs that distilled by RAG process and 10 were randomly selected from the original text data. The combination of the text distilled from the RAG process and original texts increase complexity and linguistic diversity, ensuring a robust evaluation across varied patterns and content types.


\subsubsection{Data Annotation}
Terms and relations in the training and testing datasets were manually labeled. This annotation serve as ground truth for both training and evaluation of the methods.

\subsubsection*{a) Domain-specific term annotation}

\noindent The terms are identified and classified into the top concepts from the datasets using Label Studio\textsuperscript{TM} \cite{Label}. The annotated terms were then converted into the format proposed in \cite{Liu2025}, where special symbol pairs are inserted at the start and end positions of each identified term, labelled by top concepts, to facilitate the use of the annotation results.

\subsubsection*{b) Synonym and relation annotation}

\noindent For each text containing multiple annotated terms from the previous step, the experts identified semantic relations between terms based on contextual information, forming relation triples in the format \texttt{[subject, object, relation]} as described in \tabref{tab3}. 
The \texttt{subject} and \texttt{object} are terms from the same text. 
Their roles in a triple are assigned based on their order of appearance in the text: the earlier-occurring term is designated as the subject, and the later-occurring term as the object. 

\texttt{Relation} denotes the directed relationship from subject to object. Common relation types in the training dataset are consistently labeled (e.g., "parent of": the object is a subclass or instance of the subject; "affects": the subject influences the object without direction). For other relations, the experts use the verbs or phrases originated from the text as their names (e.g., "controls" for "... solidification time controlled by the temperature ...") to enhance the generalization and induce the LLM to identify novel relations from the input text instead of only using existing relation types in the training dataset. If no relation is identified between any pair of terms in a given text, the triple is labeled as \texttt{"None"}.



\subsubsection*{c) Annotation review}
\noindent The annotation were carried out by junior domain experts with basic domain knowledge. The initial annotation underwent a comprehensive review by a domain expert to ensure terminological accuracy and consistency across the dataset. 
In total, in the training dataset, 538 domain-specific terms were identified and extracted. 96 texts containing multiple terms were annotated, from which 321 relation triples were labeled, including 7 synonym pairs. In the testing dataset, 590 terms and 481 relations from 30 paragraphs were manually annotated and validated, including
22 synonym pairs. 


\subsection{Pre-trained LLM-driven method}
In the pre-trained LLM-driven method, domain-specific terms and relations were directly extracted by the model without relying on any human-annotated examples. The LLM used in this study was ChatGPT 4.1 mini, guided by carefully engineered prompts to identify ontological elements, particularly terms and their relationships (RE). The 6 top-level domain concepts were incorporated into the prompt design to help the model focus on relevant terminology and semantic links. A training dataset was used to iteratively refine the prompts, and various formulations based on the Chain-of-Thought (CoT) \cite{Wei2022} methodology were developed and tested. The results from each prompt iteration were documented and reviewed by domain experts for validation. 

\begin{figure}[!h]
    \centering
    \includegraphics[width=0.98\linewidth]{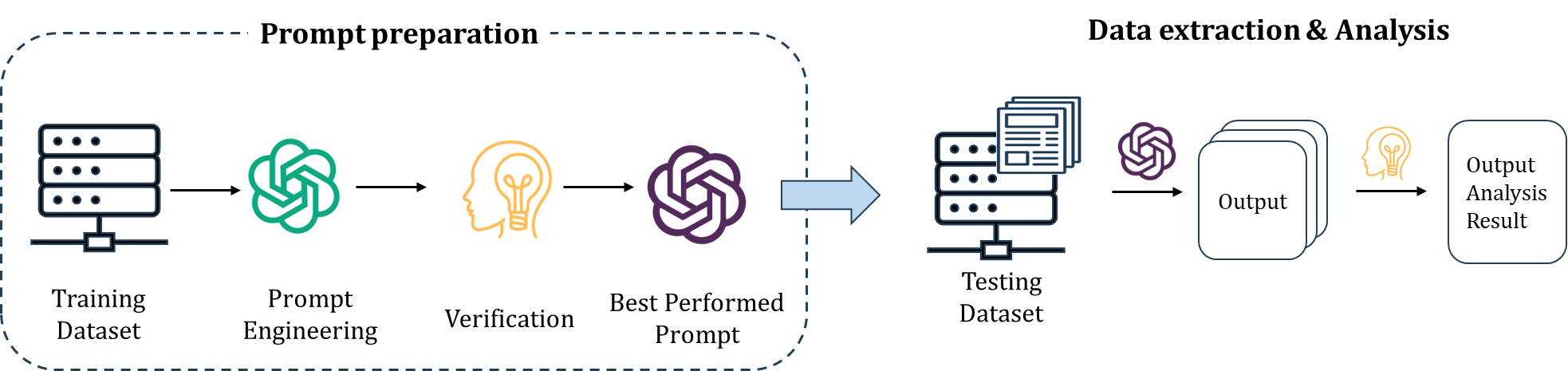}
    \caption{Pre-trained LLM extraction method process}
    \label{fig:3}
\end{figure}

The extracted terms and relations were serialized into a structured JavaScript Object Notation (JSON) format and subsequently underwent rigorous validation by domain experts in metallurgical casting. This critical review process ensured terminological precision of all terms and relations. The prompt that extracts the most terms and relations and has the highest accuracy after expert evaluation was selected. \figref{fig:3} illustrates the extraction process. 

\subsection{ICL method and fine-tuning method}

\begin{figure}[!h]
    \centering
    \includegraphics[width=0.98\linewidth]{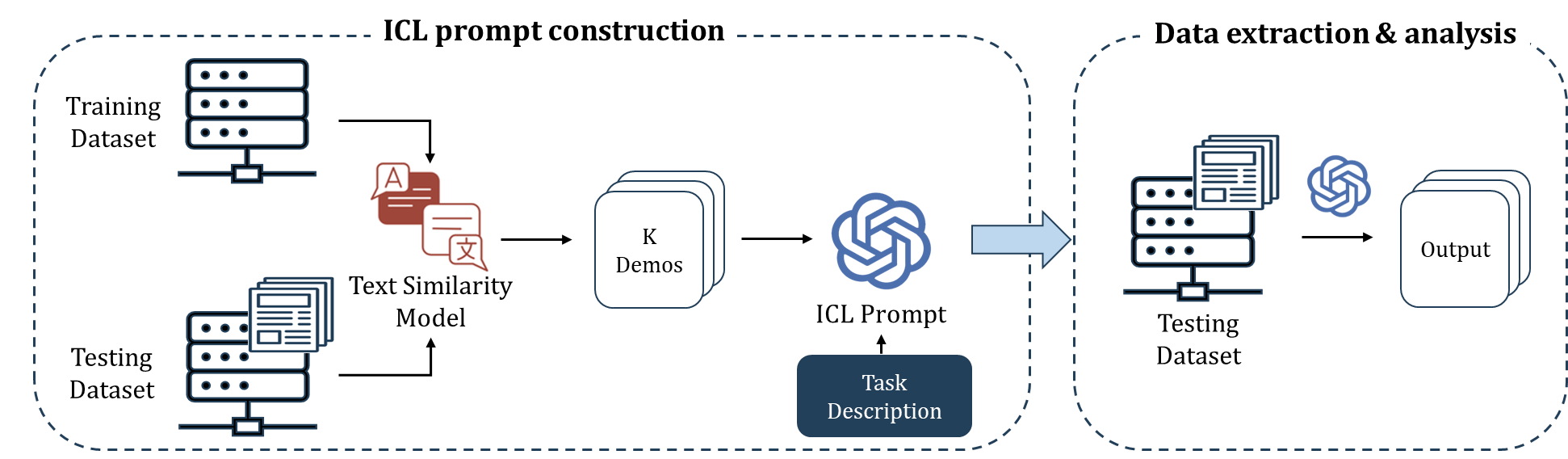}
    \caption{ICL-based method process}
    \label{fig:4}
\end{figure}

Differ from the pre-trained LLM approach, the ICL method and the fine-tuning method follow a two-step process: (1) domain-specific term extraction, and (2) synonym and relation extraction.

In step (1), LLMs are used to identify domain-specific terms from test dataset and classify them into the 6 top concepts. In the second stage, relation triples between terms are extracted, including the cross-sentence relations in the paragraphs, 
with both the subject and object drawn from the terms identified in step (1). The synonyms, e.g., "melting point" and "melting temperature", were also carefully handled in this study. For each group of synonyms, one concept was created and named using one representative term, while others were stored as synonyms for the concept. 


\subsubsection{ICL-based method}

The ICL approach uses \textit{k}-shot examples from a small training set to enhance the LLM’s ability to extract information from text without requiring model retraining. It remains effective even when the text is much longer than the examples \cite{Liu2025}. Additionally, the LLM can be guided to produce outputs in a predefined format, enabling structured analysis and evaluation. These advantages make ICL a cost-effective method for extracting structured information from long-form context, supporting domain-specific ontology construction.

As illustrated in \figref{fig:4}, the ICL method leverages examples extracted from the training dataset to guide the output format and provide the LLM with contextual references for both the term extraction and synonym and relation extraction tasks. The performance of the ICL approach depends on the semantic similarity between the input text and examples \cite{Wang2023}. Thus, the state-of-the-art text similarity model Universal AnglE Embedding (UAE)-Large-V1 \cite{li-li-2024-aoe}, which addresses saturation issues in cosine similarity was adapted to generate semantic representations for each text, enabling the calculation of cosine similarity between each training sample and test paragraph. For each test input, the \textit{k} most similar training samples are retrieved as examples.


\subsubsection*{a) Domain-specific term extraction}

\begin{table}[!h]
\caption{ICL prompt template for term extraction.}
\label{tab2}
\begin{tabular*}{\hsize}{@{\extracolsep{\fill}}llp{10cm}@{}}
\toprule
\textbf{Prompt Component} & \textbf{Content} & \textbf{Example} \\
\colrule
System Prompt & Role of LLM & You are an expert in casting and term extraction.\\
              & Task Instructions & Given a context, if the context explicitly mentions a class of casting process, use @@ and \#\# label the casting process. If the context explicitly mentions a type of material used for casting, use @@ and \$\$ ... Only add labels around terms that are mentioned in the context as related to casting. Here are some examples. \\
\colrule
User Prompt & Top \textit{k}-shot Examples & Input: Operator Training: Train operators on the proper techniques for pouring molten metal, including the use of ladles and control of pour rates ...\newline Output: Operator Training: Train operators on the proper techniques for pouring molten @@metal\$\$, including the use of @@ladles\^{}\^{} and control of @@pour rates\&\& ...\\
            & Test Input & Input: Additives play a fundamental role in metal casting by ... (entire paragraph)\newline Output:\\
\botrule
\end{tabular*}
\end{table}

\tabref{tab2} presents the ICL prompt used for casting term extraction, which comprises a system prompt and a user prompt. The system prompt defines the role of the LLM and outlines the extraction task, while the user prompt includes \textit{k}-shot examples followed by the test input.

\begin{itemize}
\item \textbf{System prompt:} It defines the role of the LLM and instructs the model to label casting-related terms according to their top concepts, using distinct symbol pairs—for example, "@@" and "\#\#" for casting process terms, and "@@" and "\texttt{||}" for product property terms. The prompt then reiterates that only terms relevant to the casting domain should be labeled and introduces the following demonstrations to guide the model.
\item \textbf{User prompt:} 
Each example includes the input text from training dataset and its corresponding annotated output with special symbol pairs. The test input text is then appended at the end of the prompt to induce the LLM to generate the labeled output.

\end{itemize}

\subsubsection*{b) Synonym and relation extraction}

\begin{table}[!h]
\caption{ICL prompt template for synonym and relation extraction.}
\label{tab3}
\begin{tabular*}{\hsize}{@{\extracolsep{\fill}}llp{10cm}@{}}
\toprule
\textbf{Prompt Component} & \textbf{Content} & \textbf{Example} \\
\colrule
System Prompt & Role of LLM & You are an expert in casting and relation extraction.\\
              & Task Instruction & Please extract relations between the listed terms from the context as triples. Do not use any term that is not listed before the context. Here are some examples. \\
\colrule
User Prompt & Top \textit{k}-shot Examples & Terms: alloy, semisolid casting, ...\newline Context: But from theoretical point, most of alloy except eutectic alloy can be cast by semisolid casting ... \newline Triples: [subject: alloy, object: semisolid casting, relation: processed by]; [subject: semisolid casting, object: magnesium, relation: processes]; ... \\
            & Test Input & Terms: Al7Si0.3Mg alloy, SSM processing, Rheocasting, ...\newline Context: The material used in the current study was an Al7Si0.3Mg alloy commonly used for SSM processing and Rheocasting ... (entire paragraph)\newline Triples:\\
\botrule
\end{tabular*}
\end{table}


Texts with multiple terms extracted in the previous step were selected for synonym and relation extraction. The demonstration with similar terms can promote performance of ICL in the relation extraction. Thus, labeled terms from each training text and extracted terms from each test paragraph were listed in their original order and prepended to the context to enhance the term information representation for demonstration retrieval. The ICL prompt for synonym and relation extraction follows a structure similar to that of the term extraction task, as illustrated in \tabref{tab3}.
\begin{itemize}


\item \textbf{System prompt:} This prompt defines the role of the LLM, instructs it to identify the semantic relations between extracted terms within a given text, and introduces the upcoming demonstrations.

\item \textbf{User prompt:} 
Each example includes the list of terms, the context from the training dataset, and the annotated triples. The extracted terms and test input text are then appended after the examples to prompt the LLM to generate output triples.
\end{itemize}

\subsubsection{Fine-tuning-based method}

The fine-tuning approach involves training two customized models with a limited dataset: one for term extraction and another for synonym and relation extraction. Unlike ICL, the fine-tuned models were trained on the entire training set to update the model parameters, requiring more time and computational resources.

\begin{figure}[!h]
    \centering
    \includegraphics[width=0.88\linewidth]{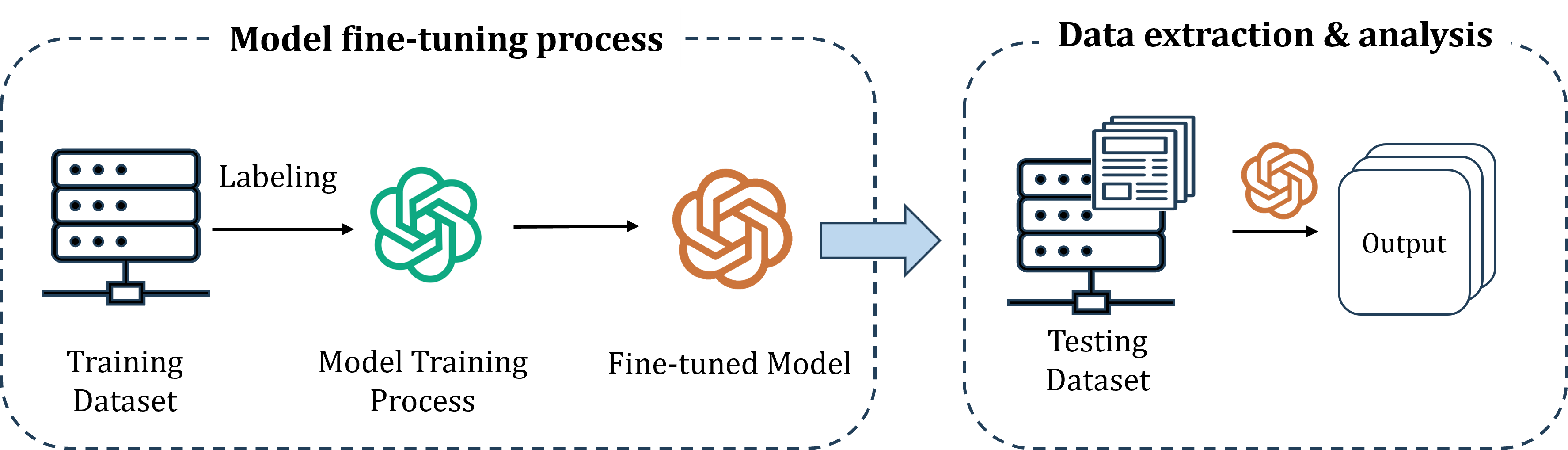}
    \caption{Fine-tuning method process}
    \label{fig:5}
\end{figure}

The workflow of the fine-tuning approach is illustrated in \figref{fig:5}.
The fine-tuning approach involves supervised fine-tuning of a chat completion LLM (OpenAI GPT-4.1-mini) using the labeled training data. The training data is organized into message that matches the model’s required format (system + user + assistant prompt). The method adapts the base model to the casting-specific information extraction tasks without needing \textit{k}-shot examples like ICL at inference time. 

Various hyperparameter configurations were tested to obtain the best-performing fine-tuned models, including the number of epochs, batch size, and learning rate (LR) multiplier. The LR multiplier, a scaling factor applied to the base learning rate defined by OpenAI, controls the step size during gradient updates. Generally, optimal values for epochs and batch size depend on the size of the training set and task complexity. For smaller training sets (fewer than 200 samples), smaller values are preferred to prevent overfitting. A larger LR multiplier may accelerate convergence but can lead to training instability. Based on our experiments, the optimal hyperparameters for fine-tuning both models on our training set are summarized in \tabref{tab4}.

\begin{table}[h]
\caption{Optimal hyperparameters for fine-tuning term extraction model and synonym \& relation extraction model}
\label{tab4}
\begin{tabular*}{\hsize}{@{\extracolsep{\fill}}lll@{}}
\toprule
\textbf{Hyperparameter} & \textbf{Term Extraction Model} & \textbf{Synonym \& Relation Extraction Model}\\
\colrule
Base model  &   gpt-4.1-mini-2025-04-14 & gpt-4.1-mini-2025-04-14\\
Epochs &   3  & 3 \\
Batch size  & 1 & 1\\
LR multiplier &  2.0 & 1.0\\
\botrule
\end{tabular*}
\end{table}


 

\subsubsection*{a) Fine-tuning domain-specific term extraction model}
The system prompt was identical to the one used in the ICL-based term extraction task but excluded demonstration introduction. The user prompt corresponded to the input text from the training dataset, while the assistant prompt provided the annotations for the input text.

In inference, a system prompt and the test input were organized in the same format as the training data and fed into the fine-tuned model for casting term recognition without any additional demonstration.

\subsubsection*{b) Fine-tuning synonym and relation extraction model}


The system prompt was identical to that used in the ICL relation extraction approach without example introduction, the user prompt included the list of terms and the original context text from the training dataset, while the assistant prompt provided the relation triple annotations for the text.

During inference, the synonym and relation extraction model received the system prompt, term list, and the text from the testing dataset. It was then able to identify both in-sentence and cross-sentence relations, including synonyms, between the listed casting terms and output the corresponding relation triples.

\section{Experiment results and discussion}


\begin{figure}[htbp]
  \centering
  \begin{subfigure}{0.48\textwidth} 
    \includegraphics[width=\linewidth]{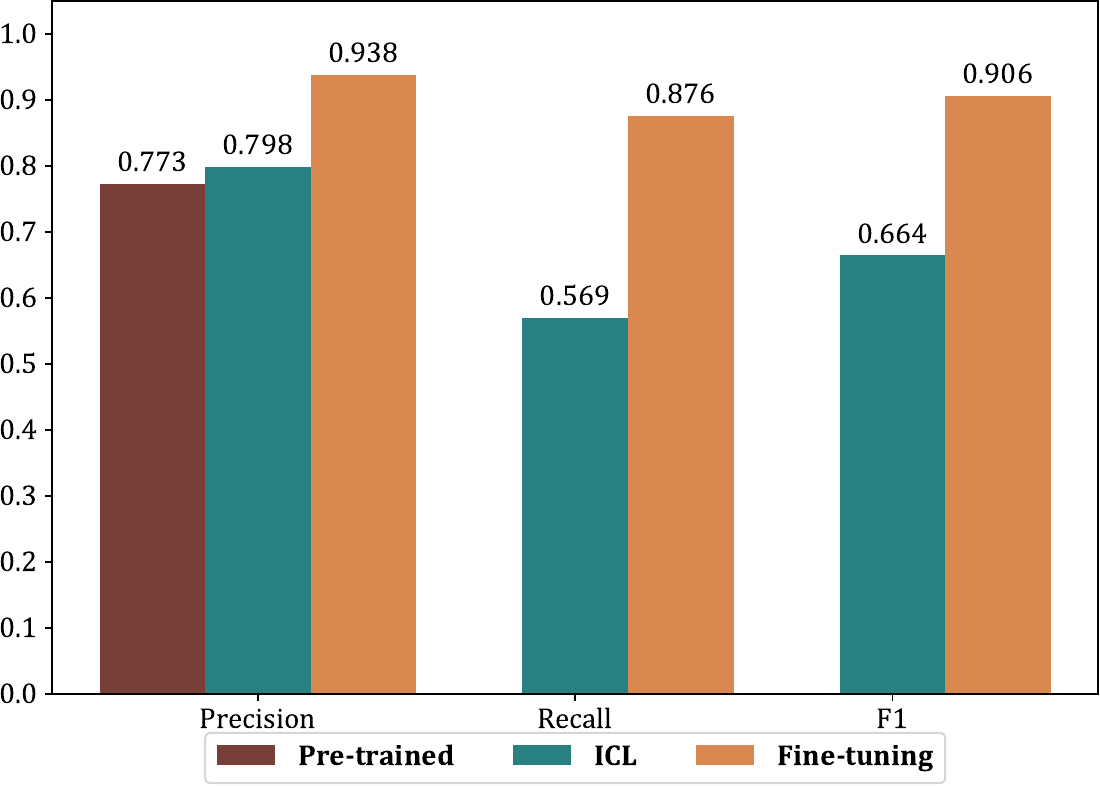}
    \caption{}
    \label{fig6:sub-a}
  \end{subfigure}
  \hspace{0.02\textwidth} 
  \begin{subfigure}{0.48\textwidth} 
    \includegraphics[width=\linewidth]{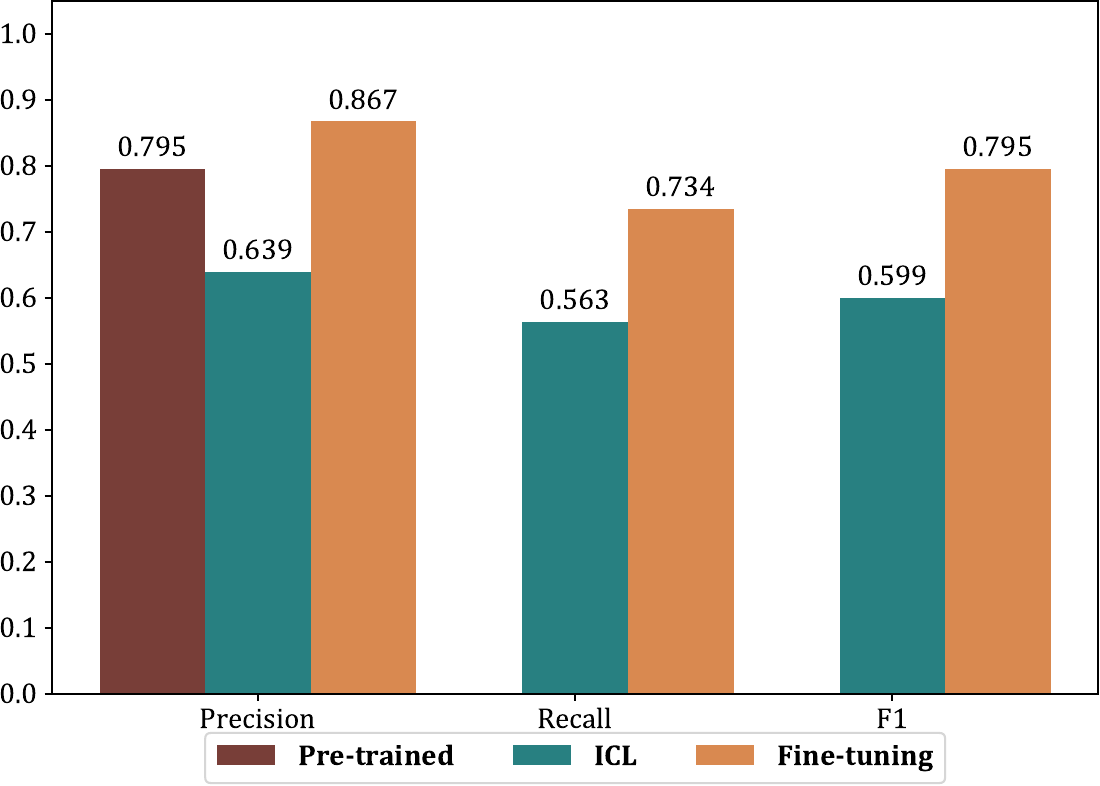}
    \caption{}
    \label{fig6:sub-b}
  \end{subfigure}
  \caption{The performances among approaches on (a) the term extraction task; (b) the synonym recognition and relation extraction task.}
  \label{fig:6}
\end{figure}
\subsection{Pre-trained LLM-driven method}
For the pre-trained LLM-driven approach,through this systematic optimisation using the training dataset, the best-performing prompt achieved approximately 80\%  accuracy for term extraction across different manufacturing domains, while relation extraction reached around 75\% accuracy.

The testing phase involved importing all the texts from the testing dataset 
into the LLM using the optimal prompt that was validated during training phase. Since the extract result is a only result and no position information, domain experts evaluated the extraction results, which gave a term extraction precision of 77. 3\% $\pm 16\%$ and a relation extraction precision of 79. 5\%$\pm 0.5\%$. The evaluation results 
for term and relation extraction precision are shown in \figref{fig:6}. In total, 142 terms and 36 relations (including different hierarchies) were extracted, which is much less than the ground truth.

Since the extraction process outputs only terms and relations without positional and context information, precision was the only available evaluation metric. Identifying false negatives was challenging due to lack of position information, as missed terms lacked traceability, preventing F1-score calculation. Additionally, the analysis relied on manual expert review, which introduced potential bias due to individual knowledge limitations. 
Without external domain-specific input, the pre-trained model depended solely on its learned knowledge, which may lead to some irrelevant or misclassified terms (e.g., rubber as a material unrelated to casting). This was partly due to high-level category definitions and the absence of detailed examples. The model also occasionally hallucinated terms, especially in longer texts, further affecting accuracy. Despite these limitations, the method offers significant efficiency over traditional manual annotation, making it a valuable starting point—particularly for junior experts working with domain-specific texts. 

\subsection{ICL-based method and fine-tuning method}
Compared with the pre-trained LLM-driven approach, both the ICL and fine-tuning methods demonstrated significantly better performance in term extraction, synonym recognition, and relation extraction tasks. A key advantage of these methods is that their outputs include either term positions or structured relation triples, enabling comprehensive evaluation using precision (P), recall (R), and the F1-score. 

The ICL approach generally benefits from larger values of \textit{k}, which controls the number of examples. In our experiment, we set \textit{k}=16 for both term and relation extraction tasks to balance token cost and performance. For each text in the testing dataset, the 16 most semantically similar texts from the training set, along with their annotated outputs, were selected as examples. 

As shown in \figref{fig6:sub-a}, the ICL method extracted a total of 421 terms—nearly three times more than the LLM-driven method. Of these, 336 matched the human-labeled ground truth, resulting in a precision of 79.8\%, a recall of 56.9\%, and an F1-score of 66.4\%.
For synonym and relation extraction, the ICL method identified 424 relation triples, with 271 correctly matched against the ground truth. This yields a precision of 63.9\%, recall of 56.3\%, and an F1-score of 59.9\% as illustrated in \figref{fig6:sub-b}.  The results  remains effective even when the text is much longer than the examples, or when interrelated terms are far apart within the text.

However, relation label inconsistency was a major limitation. Even when describing the same relation type, the ICL model output varying names such as “processed by,” “produced by,” and “used in,” despite the standardized term “processed by” being used in the training annotations. Most of these inconsistent names, as they present similar meaning, are directly captured from the input texts, revealing that the ICL method learns fewer patterns from labelled data and is crucially influenced by the expression of the input data. This inconsistency 
limited the practical use of the output.

Additionally, the ICL method showed poor performance in synonym recognition. Of the 22 synonym pairs annotated in the testing dataset, only one was successfully identified in the evaluation. This is likely due to the limited presence of synonyms in the training data, further exacerbated by their low frequency in the top \textit{k} examples. These results underscore ICL’s strong dependence on the distribution and diversity of the training data.

In contrast, the fine-tuning method achieved the highest overall performance across both information extraction tasks, as shown in \figref{fig:6}. In the term extraction task, it produced 551 terms in total, 517 of which were correct, resulting in a precision of 93.8\%, recall of 87.6\%, and an F1-score of 90.6\%. Most false negatives involved high-frequency generic terms such as “mold,” “metal,” and “alloy,” which may have been omitted due to contextual ambiguity. The false positives were primarily due to either ambiguous expressions (e.g., “matrix alloys,” “core materials”) or imprecise truncation (e.g., extracting “stirring speed” instead of “secondary stirring speed and timing”). Notably, many of the incorrect extractions appeared factually acceptable base on expert opinion, which implies that the model’s actual precision could be higher if semantic flexibility is allowed.

The fine-tuned model also extracted 407 relation triples, of which 353 were correct. This yields a precision of 86.7\%, recall of 73.4\%, and F1-score of 79.5\%. Compared to the ICL approach, the fine-tuning method showed much greater consistency in relation naming, demonstrating its strength in learning patterns from training data. One of the main sources of error, however, was relation name substitution. For instance, in the sentence: “\textit{...cast iron are often selected for their low cost and excellent fluidity, which facilitates ease of casting, though they may lack the mechanical strength...,}” the ground truth relation was labeled as \texttt{[cast iron, mechanical strength, lacks]}. The model instead generated \texttt{[cast iron, mechanical strength, has property]}, likely because the term “lacks” was absent from the training data while “has property” was more commonly seen. This type of error accounted for 17 out of the 54 false positives, illustrating the limitation imposed by the small training set. Nevertheless, these errors often retained semantic correctness, further suggesting high practical precision.
In terms of synonym recognition, the fine-tuned method outperformed ICL significantly. It identified 14 synonym pairs and correctly matched 10 of them, demonstrating a stronger ability to handle sparse synonym data in the training set.
\begin{figure}
    \centering
    \includegraphics[width=0.98\linewidth]{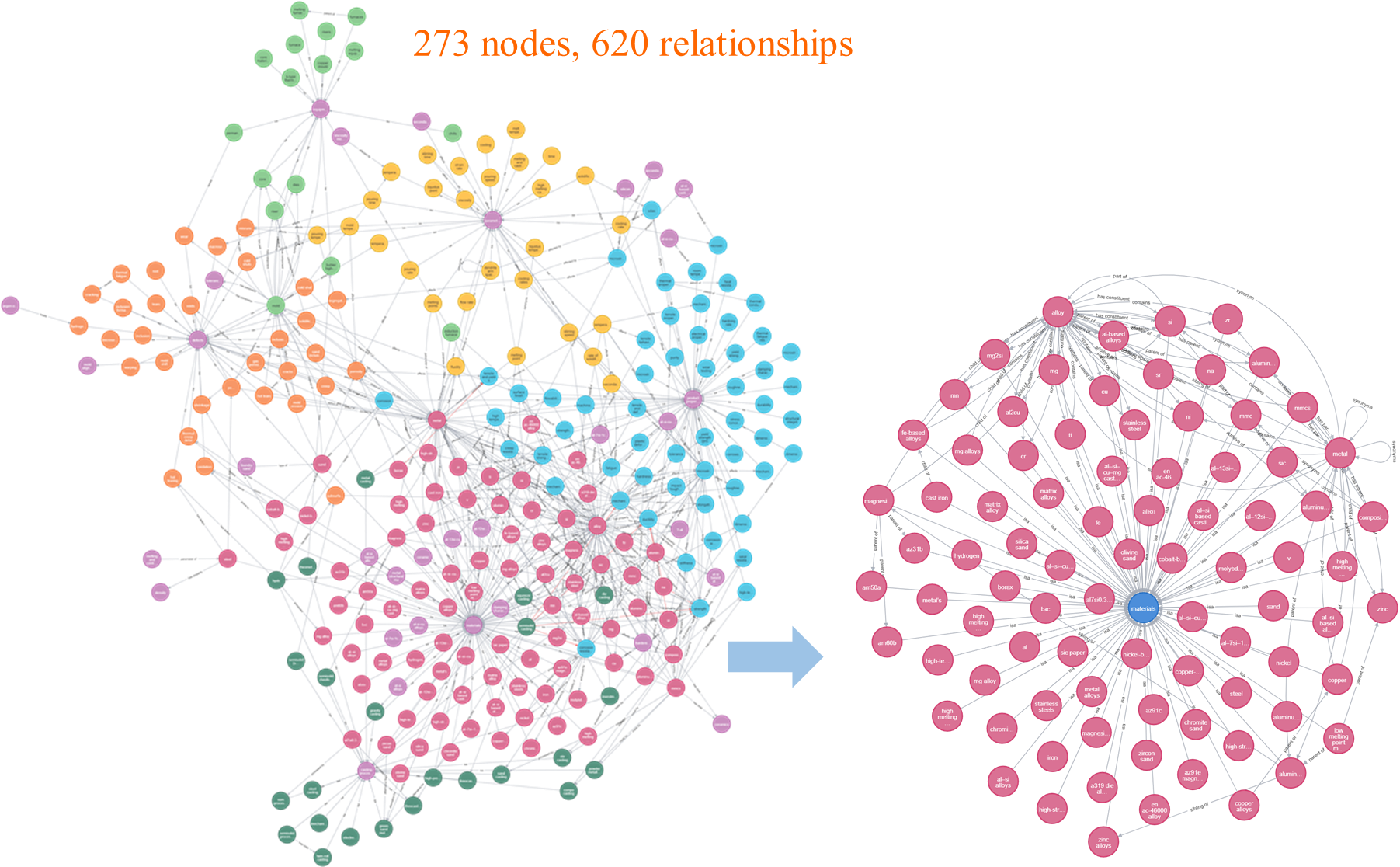}
    \caption{Overview of the constructed casting ontology and focused view of ontology subsection (materials)  }
    \label{fig:8}
\end{figure}
Overall, the fine-tuning approach proved to be the most effective among the three methods for extracting terms, synonym pairs, and relation triples from long, complex texts using limited training data. Its superior performance highlights its potential for semi-automated, domain-specific ontology construction in complex fields such as casting.

The fine-tuned method for extracting triples was used for constructing the ontology. Prior to construction, synonym terms were consolidated while the duplicate terms are deleted to form unique concepts. Additionally, the implicit "is a" relations between terms and corresponding top concepts were added to form a more complete hierarchy. 

\begin{figure}[!h]
    \centering
    \includegraphics[width=0.5\linewidth]{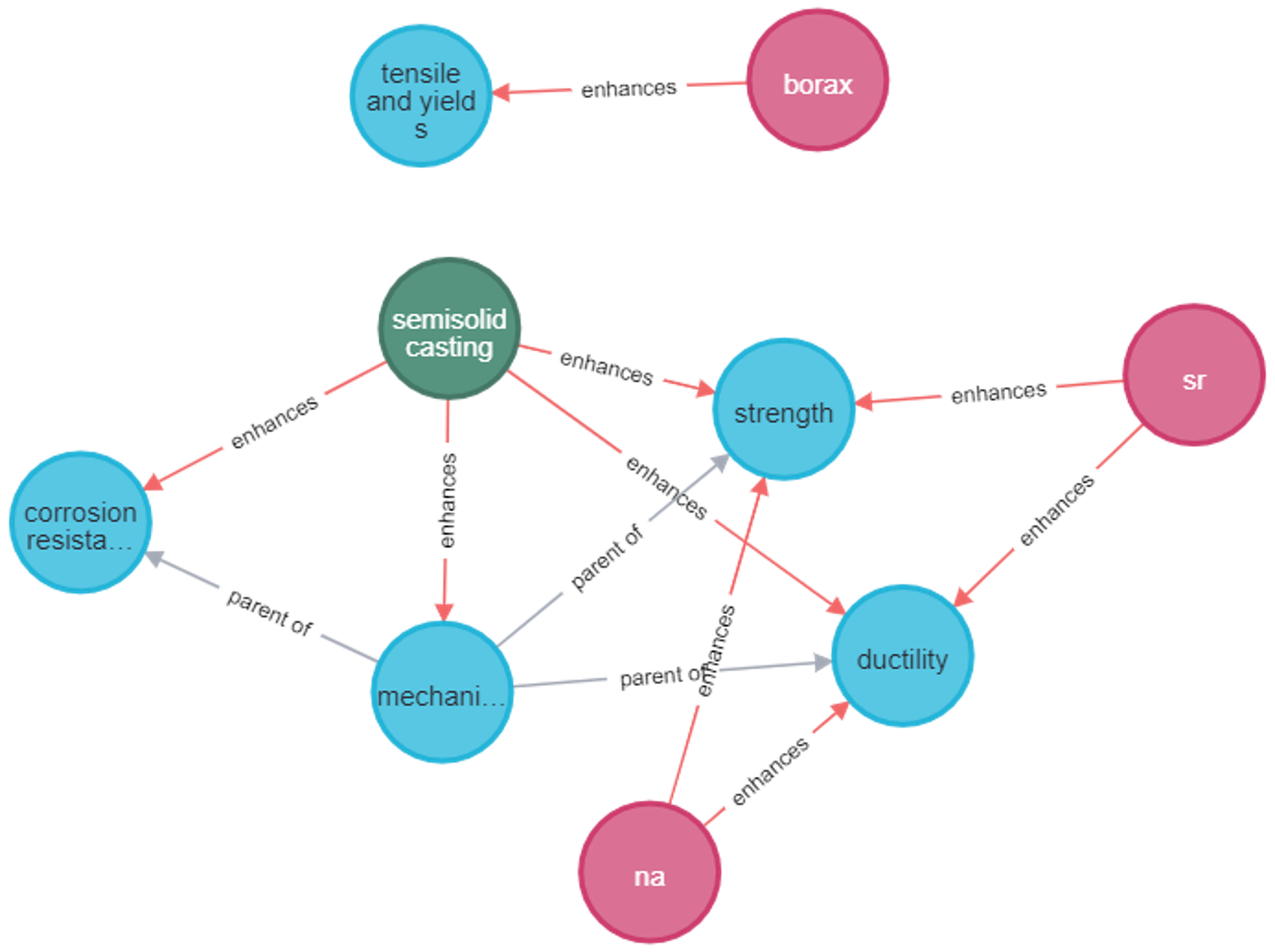}
    \caption{Detailed relation  and term illustration by neo4j}
    \label{fig:9}
\end{figure}

All remaining relation triples were evaluated by domain experts, regardless of whether the terms or relations appeared in the corresponding context. The assessment indicated that the concept precision reached 97\% and the relation precision reached 93\%. In total, 273 unique concepts and 620 validated relations were retained, meeting the quality thresholds established during expert review.\figref{fig:8} illustrates the ontology generated through Neo4j, showcasing the interconnected relationships between domain concepts and the hierarchical organisation of casting-related terminology, with different colors represent different categories, while \figref{fig:9} demonstrates a fragment of the whole ontology.

\section{Conclusion}
This study compared three LLM-based methods for extracting terms and relations to support ontology construction in the casting domain. The pre-trained model, while limited in completeness, showed strong precision and is suitable as an initial annotation aid. The ICL method was more efficient and cost-effective but suffered from inconsistent relation naming. The fine-tuned model achieved the best overall performance in both precision and recall, making it the most suitable choice for high-quality ontology development when resources allow.

While the results are promising, the study is limited by the small dataset size. Future work should evaluate these methods on larger and more diverse corpora to validate their scalability and effectiveness in real-world industrial settings.
\section*{Acknowledgements}

This research is funded by the U.S. Department of Energy's Office of Manufacturing and Energy Supply Chains (DE-EE0009726). The authors thank all the colleagues from Comptech i Skillingaryd AB, Jönköping University's Department of Material and Manufacturing and Department of Mechanical Engineering, National University of Singapore. We also acknowledge Prof. Larry Smarr and Prof. Thomas DeFanti from University of California San Diego for HyperCluster computing support of San Diego Supercomputer Center (SDSC) National Research Platform (NRP) Nautilus sponsored by the NSF (2100237, 2120019). The PhD student Xuan Liu would like to acknowledge the support from the NUS Research Scholarship. Ziyu Li PhD project was funded by the Stiftelsen för kunskaps- och kompetensutveckling (KK-Stiftelsen) via the Smart Industry Sweden initiative (project number 2020-0044)

\bibliographystyle{elsarticle-num}
\bibliography{procedia_ISM}

@inproceedings{li-li-2024-aoe,
    title = "{A}o{E}: Angle-optimized Embeddings for Semantic Textual Similarity",
    author = "Li, Xianming  and
      Li, Jing",
    editor = "Ku, Lun-Wei  and
      Martins, Andre  and
      Srikumar, Vivek",
    booktitle = "Proceedings of the 62nd Annual Meeting of the Association for Computational Linguistics (Volume 1: Long Papers)",
    month = aug,
    year = "2024",
    address = "Bangkok, Thailand",
    publisher = "Association for Computational Linguistics",
    url = "https://aclanthology.org/2024.acl-long.101/",
    doi = "10.18653/v1/2024.acl-long.101",
    pages = "1825--1839"
}

@article{Liu2025,
  title = {Knowledge extraction for additive manufacturing process via named entity recognition with LLMs},
  volume = {93},
  ISSN = {0736-5845},
  url = {http://dx.doi.org/10.1016/j.rcim.2024.102900},
  DOI = {10.1016/j.rcim.2024.102900},
  journal = {Robotics and Computer-Integrated Manufacturing},
  publisher = {Elsevier BV},
  author = {Liu,  Xuan and Erkoyuncu,  John Ahmet and Fuh,  Jerry Ying Hsi and Lu,  Wen Feng and Li,  Bingbing},
  year = {2025},
  month = jun,
  pages = {102900}
}

@article{jawad2023adoption,
  title={Adoption of knowledge-graph best development practices for scalable and optimized manufacturing processes},
  author={Jawad, MS and Dhawale, Chitra and Ramli, Azizul Azhar Bin and Mahdin, Hairulnizam},
  journal={MethodsX},
  pages={102124},
  year={2023},
  publisher={Elsevier}
}

@article{manesh2020knowledge,
  title={Knowledge management in the fourth industrial revolution: Mapping the literature and scoping future avenues},
  author={Manesh, Mohammad Fakhar and Pellegrini, Massimiliano Matteo and Marzi, Giacomo and Dabic, Marina},
  journal={IEEE Transactions on Engineering Management},
  volume={68},
  number={1},
  pages={289--300},
  year={2020},
  publisher={IEEE}
}

@article{Fan2024Unleashing,
  title={Unleashing the Potential of Large Language Models for Knowledge Graph Construction: A Practical Experiment on Incremental Sheet Forming},
  author={Fan, Haolin and Fuh, Jerry Ying Hsi and Lu, Wen Feng and Kumar, A. Senthil and Li, Bingbing},
  booktitle={Procedia Computer Science},
  volume={232},
  pages={1269--1278},
  year={2024},
  note={Presented at the International Conference on Industry 4.0 and Smart Manufacturing (ISM 2023), Lisbon, Portugal, November 22-24, 2023. *Corresponding author}
}

@misc{Label,
  title={{Label Studio}: Data labeling software},
  url={https://github.com/heartexlabs/label-studio},
  note={Open source software available from https://github.com/heartexlabs/label-studio},
  author={
    Maxim Tkachenko and
    Mikhail Malyuk and
    Andrey Holmanyuk and
    Nikolai Liubimov},
  year={2020-2022},
}

@article{Du2024,
   author = {Rick Du and Huilong An and Keyu Wang and Weidong Liu},
   month = {4},
   title = {A Short Review for Ontology Learning: Stride to Large Language Models Trend},
   url = {https://arxiv.org/pdf/2404.14991v2},
   year = {2024}
}

@article{Arp2018,
   author = {Robert Arp and Barry Smith and Andrew D. Spear},
   doi = {10.7551/mitpress/9780262527811.003.0001},
   journal = {Building Ontologies With Basic Formal Ontology},
   month = {10},
   pages = {1-26},
   publisher = {The MIT Press},
   title = {What is an Ontology?},
   url = {https://arxiv.org/pdf/1810.09171},
   year = {2018}
}

@article{Rani2017,
   author = {Monika Rani and Kumar Dhar and O P Vyas and Amit Kumar Dhar},
   doi = {10.1016/j.engappai.2017.05.006},
   journal = {Engineering Applications of Artificial Intelligence},
   pages = {108-125},
   title = {Semi-automatic terminology ontology learning based on topic modeling},
   volume = {63},
   url = {https://doi.org/10.1016/j.engappai.2017.05.006.},
   year = {2017}
}

@article{Grny2014,
   author = {Z Górny and D Wilk-Kołodziejczyk and A Smolarek-Grzyb and Andrzej Frycz and Modrzewski Krakowuniversity},
   issn = {1897-3310},
   journal = {Archives of foundry engineering},
   month = {1},
   pages = {33-36},
   title = {Supporting the Bronze CastingThrough Information Structuring Based on Ontology application},
   volume = {14},
   year = {2014}
}

@article{Borgo2007,
   author = {Stefano Borgo and Paulo Leitão},
   doi = {10.1007/978-0-387-37022-4_27},
   isbn = {978-0-387-37022-4},
   pages = {751-775},
   publisher = {Springer, Boston, MA},
   title = {Foundations for a Core Ontology of Manufacturing},
   url = {https://link.springer.com/chapter/10.1007/978-0-387-37022-4_27},
   year = {2007}
}

@article{senthil,
   author = {Senthil K Chandrasegaran and Karthik Ramani and Ram D Sriram and Imré Horváth and Alain Bernard and Ramy F Harik and Wei Gao},
   title = {The evolution, challenges, and future of knowledge representation in product design systems}
}

@article{Funk2023,
   author = {Maurice Funk and Simon Hosemann and Jean Christoph Jung and Carsten Lutz},
   isbn = {2309.09898v1},
   title = {Towards Ontology Construction with Language Models},
   year = {2023}
}

@article{Osman2022,
   author = {Mohamad Amin Osman and Shahrul Azman Mohd Noah and Saidah Saad},
   doi = {10.1109/ACCESS.2022.3163758},
   issn = {21693536},
   journal = {IEEE Access},
   pages = {43267-43283},
   publisher = {Institute of Electrical and Electronics Engineers Inc.},
   title = {Ontology-Based Knowledge Management Tools for Knowledge Sharing in Organization-A Review},
   volume = {10},
   url = {https://www.researchgate.net/publication/359658996_Ontology-Based_Knowledge_Management_Tools_for_Knowledge_Sharing_in_Organization_-_A_Review},
   year = {2022}
}

@article{Costa2013,
   author = {Ruben Costa and Celson Lima and João Sarraipa and Ricardo Jardim-Gonçalves},
   doi = {10.1007/S10845-013-0856-5/METRICS},
   issn = {15728145},
   issue = {1},
   journal = {Journal of Intelligent Manufacturing},
   month = {12},
   pages = {263-282},
   publisher = {Springer New York LLC},
   title = {Facilitating knowledge sharing and reuse in building and construction domain: An ontology-based approach},
   volume = {27},
   url = {https://link.springer.com/article/10.1007/s10845-013-0856-5},
   year = {2013}
}

@inbook{inbook,
author = {Guarino, Nicola},
year = {1995},
month = {01},
pages = {25-32.},
title = {Ontologies and knowledge bases: towards a terminological clarification}
}

@inproceedings{Kitamura2003,
   author = {Yoshinobu Kitamura and Riichiro Mizoguchi},
   doi = {10.1016/S0957-4174(02)00138-0},
   issn = {09574174},
   issue = {2},
   booktitle = {Expert Systems with Applications},
   month = {2},
   pages = {153-166},
   title = {Ontology-based description of functional design knowledge and its use in a functional way server},
   volume = {24},
   year = {2003}
}

@inbook{Bittner2005,
   author = {Thomas Bittner and Maureen Donnelly and Stephan Winter},
   doi = {10.1201/9781420036282.pt3},
   isbn = {9781420036282},
   booktitle = {Large-Scale 3D Data Integration: Challenges and Opportunities},
   month = {1},
   pages = {139-160},
   publisher = {CRC Press},
   title = {Ontology and semantic interoperability},
   year = {2005}
}

@article{Sarkar2019,
   author = {Arkopaul Sarkar and Dušan Šormaz},
   doi = {10.1016/J.PROMFG.2020.01.244},
   issn = {2351-9789},
   journal = {Procedia Manufacturing},
   month = {1},
   pages = {1889-1898},
   publisher = {Elsevier},
   title = {Ontology Model for Process Level Capabilities of Manufacturing Resources},
   volume = {39},
   url = {https://www.sciencedirect.com/science/article/pii/S2351978920303012?via%3Dihub},
   year = {2019}
}

@inbook{MartinezLastra2008,
   author = {Jose L. Martinez Lastra and Ivan M. Delamer},
   doi = {10.1007/978-3-540-89784-2_11},
   isbn = {978-3-540-89784-2},
   issn = {1611-3349},
   booktitle = {Advances in Web Semantics I},
   pages = {276-289},
   publisher = {Springer, Berlin, Heidelberg},
   title = {Ontologies for Production Automation},
   volume = {4891 LNCS},
   url = {https://link.springer.com/chapter/10.1007/978-3-540-89784-2_11},
   year = {2008}
}

@article{Cao2022,
   author = {Jianpeng Cao and Edlira Vakaj and Ranjith K. Soman and Daniel M. Hall},
   doi = {10.1016/J.AUTCON.2022.104277},
   issn = {0926-5805},
   journal = {Automation in Construction},
   month = {7},
   pages = {104277},
   publisher = {Elsevier},
   title = {Ontology-based manufacturability analysis automation for industrialized construction},
   volume = {139},
   url = {https://www.sciencedirect.com/science/article/pii/S0926580522001509},
   year = {2022}
}

@misc{datatrasferforcasting,
   author = {DFA},
   title = {Transform Brownfield casting processes into Industry 4.0 compliant data space - DFA},
   url = {https://digitalfactoryalliance.eu/transform-brownfield-casting-processes-into-industry-4-0-compliant/-data-space/}
}

@misc{diecastingtrasnfer,
   author = {Dileep Yadav and Christian Kleeberg and U CHANG ENG},
   title = {foundry-planet.com - B2B Portal: “DIE-CASTING 4.0”: Driving the Digital Transformation in High Pressure Die Casting},
   url = {https://www.foundry-planet.com/d/rgu-die-casting-40-driving-the-digital-transformation-in-high-/pressure-die-casting/}
}

@article{Joachimiak2024,
   author = {Marcin P. Joachimiak and Mark A. Miller and J. Harry Caufield and Ryan Ly and Nomi L. Harris and Andrew Tritt and Christopher J. Mungall and Kristofer E. Bouchard},
   month = {4},
   title = {The Artificial Intelligence Ontology: LLM-assisted construction of AI concept hierarchies},
   url = {https://arxiv.org/pdf/2404.03044},
   year = {2024}
}

@techReport{Zhang2024,
   author = {Shiyao Zhang and Yuji Dong and Terry R Payne and Jie Zhang and Yichuan Zhang},
   title = {Large Language Model Assisted Multi-Agent Dialogue for Ontology Alignment: Extended Abstract},
   url = {https://www.researchgate.net/publication/378332970},
   year = {2024}
}

@article{tan2024,
   author = {Zhen Tan and Dawei Li and Song Wang and Alimohammad Beigi and Bohan Jiang and Amrita Bhattacharjee and Mansooreh Karami and Jundong Li and Lu Cheng and Huan Liu},
   pages = {930-957},
   title = {Large Language Models for Data Annotation and Synthesis: A Survey}
}

@article{Liao2025,
   author = {Xingming Liao and Chong Chen and Zhuowei Wang and Ying Liu and Tao Wang and Lianglun Cheng},
   doi = {10.1016/J.AEI.2025.103134},
   issn = {1474-0346},
   journal = {Advanced Engineering Informatics},
   month = {5},
   pages = {103134},
   publisher = {Elsevier},
   title = {Large language model assisted fine-grained knowledge graph construction for robotic fault diagnosis},
   volume = {65},
   url = {https://www-sciencedirect-com.proxy.library.ju.se/science/article/pii/S1474034625000278},
   year = {2025}
}

@article{Wang2023,
   author = {Shuhe Wang and Xiaofei Sun and Xiaoya Li and Rongbin Ouyang and Fei Wu and Tianwei Zhang and Jiwei Li and Guoyin Wang},
   month = {4},
   title = {GPT-NER: Named Entity Recognition via Large Language Models},
   url = {https://arxiv.org/pdf/2304.10428},
   year = {2023}
}

@article{Wei2022,
   author = {Jason Wei and Xuezhi Wang and Dale Schuurmans and Maarten Bosma and Brian Ichter and Fei Xia and Ed H. Chi and Quoc V. Le and Denny Zhou},
   isbn = {9781713871088},
   issn = {10495258},
   journal = {Advances in Neural Information Processing Systems},
   month = {1},
   publisher = {Neural information processing systems foundation},
   title = {Chain-of-Thought Prompting Elicits Reasoning in Large Language Models},
   volume = {35},
   url = {https://arxiv.org/pdf/2201.11903},
   year = {2022}
}




\clearpage

\normalMode

\end{document}